%% file: main.tex
\documentclass[sigconf]{acmart}

\AtBeginDocument{%
  }

\setcopyright{acmlicensed}
\copyrightyear{2022}
\acmYear{2022}

\acmConference[CIKM '22] {Proceedings of the 31st ACM International Conference on Information and Knowledge Management}{October 17--21, 2022}{Atlanta, GA, USA.}
\acmBooktitle{Proceedings of the 31st ACM International Conference on Information and Knowledge Management (CIKM '22), October 17--21, 2022, Atlanta, GA, USA}
\acmPrice{15.00}
\acmISBN{978-1-4503-9236-5/22/10}
\acmDOI{10.1145/3511808.3557246}
\usepackage{makecell}
\usepackage{graphicx}
\usepackage{float} 
\usepackage{subfigure}
\usepackage{stfloats}
\usepackage{multirow}
\usepackage{color}
\usepackage{xr}

\begin{document}

\title{AutoQGS: Auto-Prompt for Low-Resource Knowledge-based Question Generation from SPARQL}

\input{00-00-authors}

\input{00-01-abstract}

\ccsdesc[500]{Computing methodologies~Natural language generation}

\keywords{Knowledge Graph, Complex Question Generation, Low Resource}

\maketitle

\input{01-introduction}

\input{02-relatedwork}
\input{03-approach}
\input{04-experiments}

\input{05-conclusion}
\input{06-acks}

\input{07-appendix}


\bibliographystyle{ACM-Reference-Format}
\bibliography{qg}


\end{document}

%% file: 00-00-authors.tex
\author{Guanming Xiong}
\affiliation{
  \institution{Peking University}
  \state{Beijing}
  \country{China}
}
\email{gm\_xiong@pku.edu.cn}

\author{Junwei Bao}
\authornote{Corresponding author.}
\affiliation{
  \institution{JD AI Research}
  \state{Beijing}
  \country{China}
}
\email{baojunwei001@gmail.com}

\author{Wen Zhao}
\affiliation{
  \institution{Peking University}
  \state{Beijing}
  \country{China}
}
\email{zhaowen@pku.edu.cn}

\author{Youzheng Wu}
\affiliation{
  \institution{JD AI Research}
  \state{Beijing}
  \country{China}
}
\email{wuyouzheng1@jd.com}

\author{Xiaodong He}
\affiliation{
  \institution{JD AI Research}
  \state{Beijing}
  \country{China}
}
\email{hexiaodong@jd.com}


\renewcommand{\shortauthors}{Guanming Xiong et al.}

%% file: 00-01-abstract.tex
\begin{abstract}

This study investigates the task of knowledge-based question generation (KBQG).
Conventional KBQG works generated questions from fact triples in the knowledge graph, which could not express complex operations like aggregation and comparison in SPARQL.
Moreover, due to the costly annotation of large-scale SPARQL-question pairs, KBQG from SPARQL under low-resource scenarios urgently needs to be explored.
Recently, since the generative pre-trained language models (PLMs) typically trained in natural language (NL)-to-NL paradigm have been proven effective for low-resource generation, e.g., T5 and BART, how to effectively utilize them to generate NL-question from non-NL SPARQL is challenging. 
To address these challenges, AutoQGS, an auto-prompt approach for low-resource KBQG from SPARQL, is proposed. 
Firstly, we put forward to generate questions directly from SPARQL for the KBQG task to handle complex operations. 
Secondly, we propose an auto-prompter trained on large-scale unsupervised data to rephrase SPARQL into NL description, smoothing the low-resource transformation from non-NL SPARQL to NL question with PLMs.
Experimental results on the WebQuestionsSP, ComlexWebQuestions 1.1, and PathQuestions show that our model achieves state-of-the-art performance, especially in low-resource settings. 
Furthermore, a corpus of 330k factoid complex question-SPARQL pairs is generated for further KBQG research.\footnote{Code and data are available at: \url{https://github.com/JimXiongGM/AutoQGS}}

\end{abstract}

%% file: 01-introduction.tex
\section{Introduction} 

\begin{figure*}
    \centering
    \includegraphics[width=0.95\textwidth,page=1]{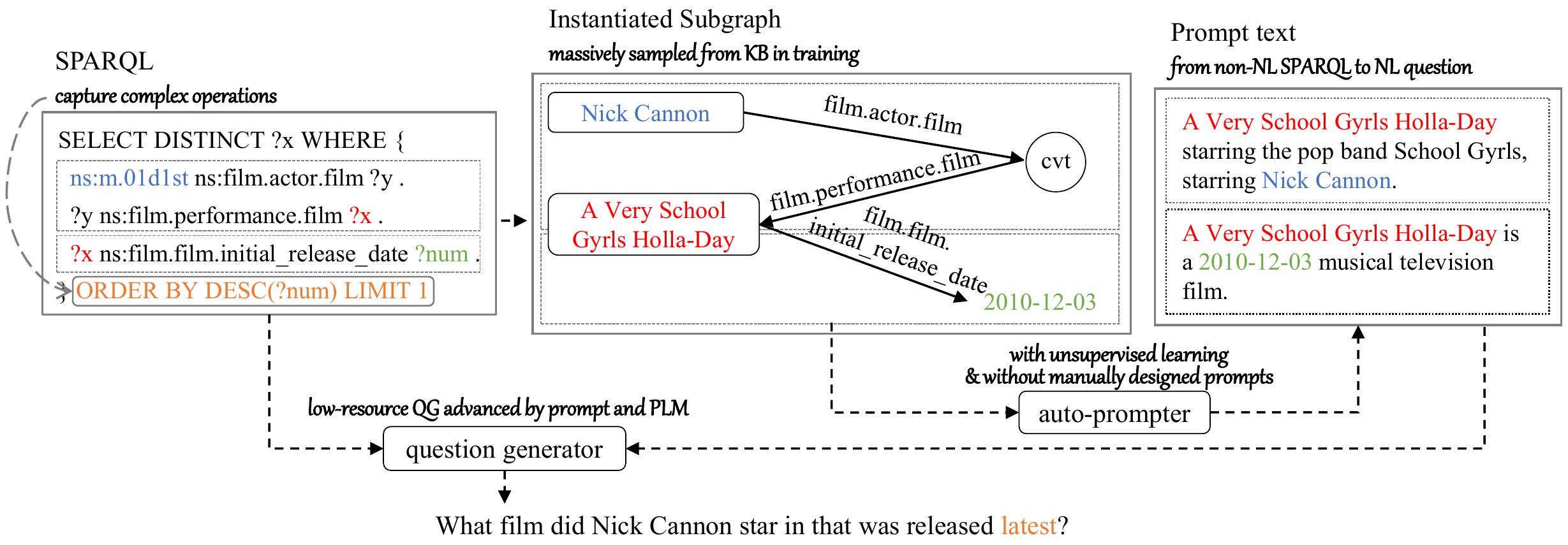}
    \caption{Model overview.}
    \label{fig:model_overview}
\end{figure*}

Knowledge-Based Question Generation (KBQG) is a task that aims to generate natural language questions given a knowledge base (KB). 
KBQG has a wide range of applications and has won wide attention in academia and industry. 
In this paper, we use a specific knowledge graph (KG), Freebase\cite{bollacker2008freebase}, as KB. 
Recent works mainly adopt sequence-to-sequence neural networks to generate questions given a Resource Description Framework \cite{miller1998introduction, Eric2001Introduction} (RDF) graph that is a directed graph composed of triple statements in a knowledge graph.
For a single relation graph, a series of works \cite{elsahar_zero-shot_2018, liu_generating_2019, bi_knowledge-enriched_2020} enriched the input with auxiliary information, equipped the decoder with copy mechanisms, and devised delicate models to improve the fluency or the semantic accuracy of generated questions. 
For star-like graph, \cite{kumar_difficulty-controllable_2019} used Transformer encoder while \cite{chen_toward_2022} applied a bidirectional Graph2Seq model to encode the graph structure.
Very recently, JointGT \cite{ke_jointgt_2021} adopted the modern pre-trained language model BART \cite{lewis_bart_2020} to generate questions, achieving state-of-the-art performance.

However, KBQG still faces three major challenges:

\paragraph{1. Absence of complex operations}
The existing KBQG methods mainly generated questions from an RDF graph. However, RDF is a standard model for data interchange on the Web, it's oriented to describe resources not to express constraints on top of them, such as aggregation, superlative, and comparative questions. Compared with the RDF graph, SPARQL has a complete semantic representation that covers all the question types mentioned above. How to generate questions based on SPARQL is a challenge.

\paragraph{2. Low resources}
For expressing the predicates in KG to NL form, traditional supervised methods tended to annotate large-scale SPARQL-question pairs. However, the labor cost is enormous. Besides, KG may contain complex schemas, such as  Compound Value Type\footnote{A Compound Value Type is a Type within Freebase which is used to represent data where each entry consists of multiple fields. See \url{https://developers.google.com/freebase/guide/basic_concepts}} (CVT) in Freebase. Each CVT combination has a different meaning, hence it is difficult for annotations to cover all combinations. How to generate questions without sufficient resources is still under-explored to date.

\paragraph{3. Efficient generation}
The advanced generative pre-trained language models (PLMs) have been proven effective in natural language generation (NLG) tasks \cite{lewis_bart_2020, chen_few-shot_2020, ke_jointgt_2021}. Nonetheless, PLMs were trained in NL-to-NL paradigm, but the SPARQL expression is different from the NL form. Therefore, how to leverage the strengths of PLMs to generate high-quality questions matters a lot.

To address the challenges mentioned above, we propose AutoQGS, an auto-prompt approach for low-resource KBQG from SPARQL. Figure \ref{fig:model_overview} shows the overall process.
Firstly, we incorporate SPARQL expression directly as input, which retains the original semantics. 
Secondly, we propose a model, auto-prompter, to rephrase the SPARQL to the corresponding NL description, named prompt text. 
Auto-prompter combines the strengths of distant supervision \cite{mintz-etal-2009-distant} and the strong generative ability of PLMs. 
Specifically, the training data of auto-prompter are subgraphs that could be massively sampled from KB, and the target prompt text is collected from large-scale corpus as well. 
Lastly, we explore an efficient question generation method in low-resource scenarios. 
Our model significantly outperforms existing state-of-the-art baselines by a large margin, especially in low-resource settings.

The main contributions of this paper are summarized as follows. 
\begin{itemize}
    \item Put forward to generate questions directly from SPARQL for KBQG task to handle complex operations.
    \item Propose AutoQGS, an approach to rephrase RDF graph to prompt text, and generate questions from SPARQL and prompt text.
    \item Conduct extensive experiments on two datasets, and the results show that AutoQGS improve the performance observably.
\end{itemize}

%% file: 02-relatedwork.tex
\section{Related Work} 

KBQG has come a long way in the past decades. In the early time, many works generate questions in template-based approaches. 
\cite{song_question_2017} proposed an unsupervised system to collect questions by a search engine from a small number of template-based questions.
\cite{Seyler2015Generating, seyler_knowledge_2017} collect structured triple-pattern query from seed question and use a template-based method to verbalize the structured query. 
However, template-based approaches lack flexibility, and the annotation cost is high.

Recent works for KBQG are mainly based on sequence-to-sequence neural networks given a set of subgraphs from knowledge graph (KG) . 
\cite{serban_generating_2016} first used a neural network for encoding KG fact triples into natural language questions and generated the 30M Factoid Question-Answer datasets. However, it was trained on a mass of fact-question pairs, which is challenging to collect. 
\cite{reddy_generating_2017} proposed an RNN-based model to generate simple questions and corresponding answers by converting all the KG entities to a set of keywords.
\cite{Bao2018Table} and \cite{Bao2019TextGeneration} developed a flexible copying mechanism to alleviate the rare words problem.
For unseen predicates and entity types problem, \cite{elsahar_zero-shot_2018} collected textual contexts in the Wikipedia as auxiliary information and adopted a part-of-speech copy action mechanism to generate questions. 
However, \cite{liu_generating_2019} thought the textual contexts were noisy or even wrong. They presented a complicated model that integrates diversified off-the-shelf contexts and devised an answer-aware loss to make sure the generated questions are associated with a definitive answer. 
Based on the Transformer \cite{vaswani_attention_2017}, \cite{kumar_difficulty-controllable_2019} proposed an end-to-end neural network-based method for generating complex multi-hop and difficulty-controllable questions over a subgraph in KG. 
\cite{bi_knowledge-enriched_2020} focus on semantic drift problem. They proposed to incorporate auxiliary information and word types in generated questions, make the decoder output conditioned on these types, and design a DPT-based evaluator to encourage question structural conformity in a reinforcement learning framework. 
Instead of using a set of KG triples, \cite{chen_toward_2022} proposed to apply a bidirectional Graph2Seq model to encode the KG subgraph and target answers, and then generate questions with a node-level coping mechanism.
\cite{ke_jointgt_2021} proposed a pre-trained model called JointGT for KG-to-text generation tasks. They added a structure-aware semantic aggregation module in BART to model the structure of input graphs. They then pre-trained the model in a large-scale corpus, and fine-tuned in question generation task. 

In comparison, our approach utilizes advanced generative pre-trained language models to generate questions from SPARQL, rather than either make templates to rephrase questions or generate questions from an KG subgraph.

%% file: 03-approach.tex
\section{Approach}

\subsection{Problem Formulation}
This study investigates the task of knowledge-based question generation (KBQG).
Conventional KBQG works generated questions from fact triples in the knowledge graph, which could not express complex operations.
A SPARQL expression is of complete syntax, which is capable of fully formalizing questions with complex operations\footnote{Complex operations are defined as functions beyond the KG predicates.}, e.g., aggregation(\textsf{COUNT}), comparative(\textsf{<,>,<=,>=}), and superlative(\textsf{ORDER BY ?x LIMIT 1}).
Moreover, due to the costly annotation of large-scale SPARQL-question pairs, KBQG from SPARQL under low-resource scenarios urgently needs to be explored.
In this paper, we propose to \textbf{generate question directly from SPARQL for low-resource KBQG} task. 
Given an executable SPARQL expression $S$ and a knowledge graph $\mathcal{K}$, our goal is to generate a natural language (NL) question $Q$ that is consistent with the SPARQL.
Given the above definitions, the task can be formalized as learning the distribution $p(Q|S, \mathcal{K})$.

\subsection{Model Overview}
Recently, the generative pre-trained language models (PLMs) typically trained in NL-to-NL paradigm have been proven effective for low-resource generation, e.g., T5\cite{2020t5} and BART\cite{lewis_bart_2020}. 
However, generating questions directly from non-NL SPARQL is not friendly to the generative PLMs.
In this paper, we propose AutoQGS, an auto-prompt approach which rephrases SPARQL to NL text automatically, smoothing the transformation from non-NL SPARQL to NL question. 
The overall process of AutoQGS is shown in Figure \ref{fig:model_overview}.
Specifically, AutoQGS consists of two procedures, (1) \textbf{auto-prompt} from SPARQL to NL text, and (2) \textbf{question generation} (QG) based on SPARQL and NL prompt text. 

Formally, auto-prompt aims to generate prompt text $T$ from SPARQL $S$.

The process of auto-prompt can be formalized as
\begin{equation}
	\label{eq:ap_overview}
	T = \mathtt{AP}(S, \mathcal{K}; \Theta)
\end{equation}

Afterwards, $S$ and $T$ are concatenated as input\footnote{We replace the machine identifiers of the topic entities in SPARQL with their surface names, but for brevity, we still use the symbol $S$.} of a question generator to produce a question $\hat{Q}$. 
The procedure of question generation can be formalized as 
\begin{equation}
\label{eq:qg_overview}
\begin{aligned}
	Q = \mathtt{QG}(S, T; \Phi)
\end{aligned}
\end{equation}
We will introduce the auto-prompt $\mathtt{AP}(\cdot)$ and the question generation $\mathtt{QG}(\cdot)$ procedures in Section \ref{sec:ap} and Section \ref{sec:qg}, respectively.

\begin{figure*}
  \centering
  \includegraphics[width=1\textwidth,page=2]{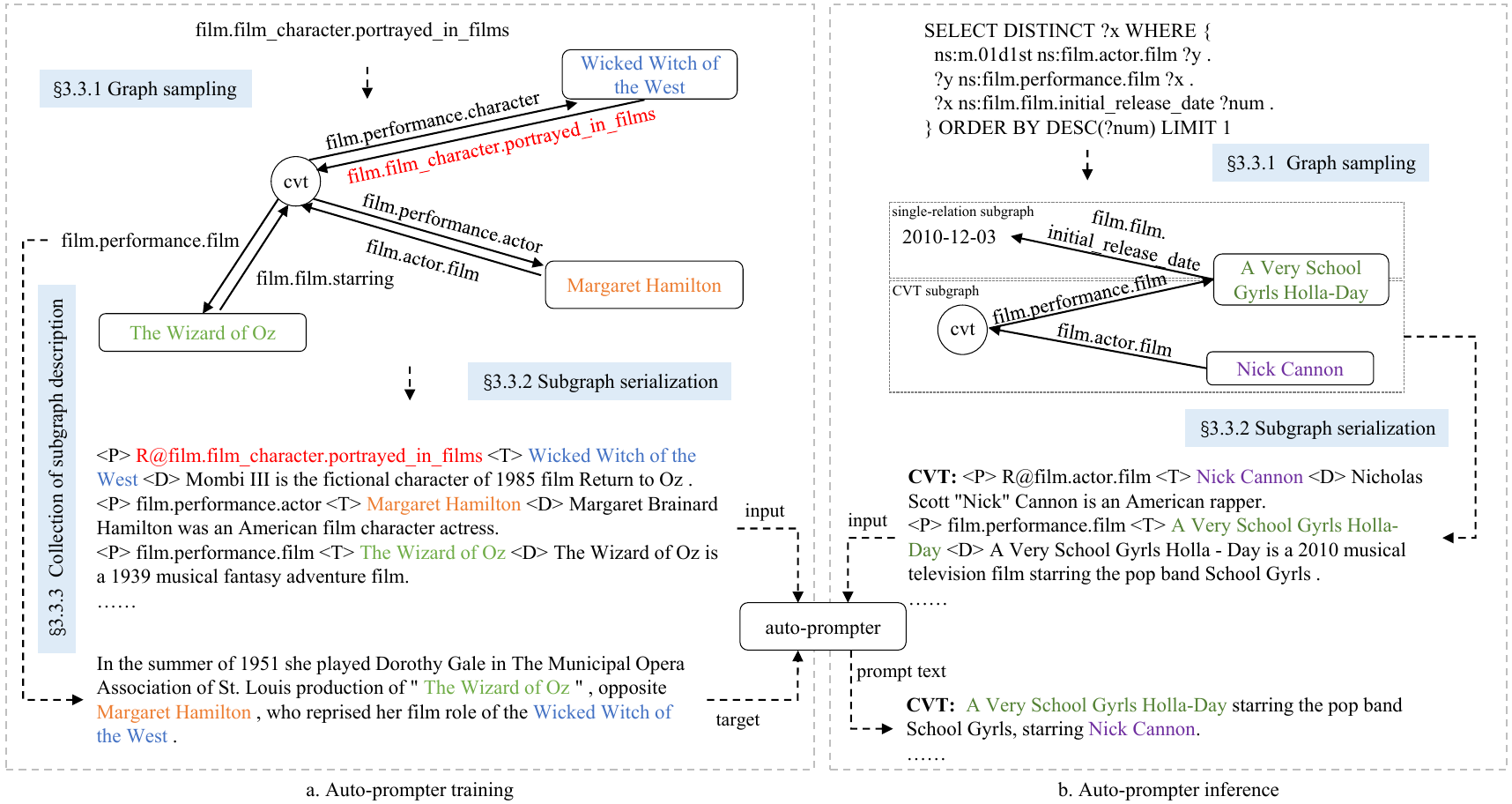}
  \caption{Overview of auto-prompt training and inference procedures. \protect\footnotemark}
  \label{fig:ap_overview}
\end{figure*}
\footnotetext{For brevity, we omit the prefix: \textsf{PREFIX ns: <http://rdf.freebase.com/ns/>}.}

\subsection{Auto-prompt}\label{sec:ap}

Auto-prompt $\mathtt{AP}(\cdot)$ generates prompt text $T$ from SPARQL $S$.
Figure~\ref{fig:ap_overview} shows the overview of auto-prompt.
Given a SPARQL $S$, we first execute it based on $\mathcal{K}$ and obtain an instantiated subgraph $G$ via subgraph sampling defined in Section \ref{subgraph_sampling}.
$G$ can be decomposed into a set of atomic subgraphs. Each $g \in G$ is serialized as input of an auto-prompter and converted into text.
They are then concatenated to construct the prompt text $T$.
Specifically, this procedure is formalized as follows:
\begin{equation}
	\label{eq:ap_detail}
	T = \mathtt{CONCAT}(\{ \mathtt{BeamSearch_{Top1}}(p_{\Theta}(d_{i}|d_{<i},g)) | g \in G\})
\end{equation}
\begin{equation}
	\label{eq:ap_g}
	G = f(S, \mathcal{K})
\end{equation}
where function $\mathtt{CONCAT}(\cdot)$ means concatenating all the elements, $\mathtt{BeamSearch_{Top1}}(\cdot)$ is the generation process with beam search (we keep the one with the highest score), $\Theta$ is the learnable parameters of auto-prompter, $d_i$ is the $i$-th predicted token, and function $f(\cdot)$ indicates the graph sampling process.
In this work, the auto-prompter supports both star and chain topologies, and the prompt text will vary as the input entities and their information change. 
By contrast, previous works only save the hard matched results for single-relation predicate (e.g. ``\textsf{is birthplace of}'' for ``\textsf{person/place\_of\_birth}'').

\subsubsection{Subgraph Sampling}\label{subgraph_sampling}
Subgraph sampling is a process that aims to find a KG subgraph that can instantiate a given SPARQL based on $\mathcal{K}$.
Firstly, we define an atomic subgraph $g$, $type(g) \in \{ CVT, Single \}$ in Freebase, where $type(\cdot)$ is a function that identifies the type of an atomic subgraph.
$CVT$ is a Compound Value Type (CVT) subgraph and $Single$ is a single-relation one.
A CVT subgraph consist of a central CVT node and its corresponding one-hop edges.
A single-relation subgraph can be formalized to a (subject, predicate, object) triple.
Specifically, we use Virtuoso to store Freebase.
Following Google's instruction\footnote{\url{https://github.com/google/Freebase-wikidata-converter}}, we classify the type of predicates in Freebase into single-relation and CVT. 
The difference is that the tail node of a CVT predicate is a CVT node and the entire CVT graph expresses an event, as the predicate ``\textsf{film.film\_character.portrayed\_in\_films}'' shown in Figure \ref{fig:ap_overview} a. 
The single-relation one links two named entities, which means a fact.

In the training stage, given a predicate, we sample a set of corresponding atomic subgraphs in the database. 
For single-relation predicate, we simply construct ``\textsf{SELECT ?s ?o WHERE \{?s [predicate] ?o.\}}'' to query KG and save all \textsf{(subject name, predicate, object name)} triples returned. 
For each CVT predicate, we sample a set of nodes that are the tails of the predicate, then instantiate a graph center at the node for each one in the set. 
As shown in Figure \ref{fig:ap_overview} a, the instantiated CVT graph consists of both inside and outside edges in one-hop connection. 
For example, in order to sample the graphs of predicate ``\textsf{film.film\_character.portrayed \_in\_films}'', we firstly construct a query, ``\textsf{SELECT ?x WHERE \{?y film.film\_character.portrayed\_in\_films ?x.\}}'', to get corresponding CVT nodes, and then for each node (e.g. \textsf{m.0gxrhxd}), we construct another query, ``\textsf{SELECT ?e1 ?p\_in ?e2 ?p\_out WHERE \{?e1 ?p\_in m.0gxrhxd. m.0gxrhxd ?p\_out ?e2.\}}'', to instantiate the CVT graph. 
Due to limited resources, instead of covering the entire predicates in Freebase, we selected a subset of predicates involved in the two datasets mentioned in Section \ref{dataset_preprocess}.

In the inference stage, given a SPARQL expression $S$, we search all variables (e.g. replacing ``\textsf{?x}'' with ``\textsf{?y ?x ?num}'' in the SPARQL) in KG and sample one from the results to instantiate a complete graph $G$. 
As shown in Figure \ref{fig:ap_overview} b, a SPARQL may consist of both single-relation and CVT subgraphs.

\subsubsection{Subgraph Serialization}\label{subgraph_serialization}

To explore how to mine the general statements more effectively for different entities with the same predicate, we propose two strategies of serialization.

\paragraph{Entity name} In this setup, we serialize the subgraph with the original entity name. As we know, Wikipedia is used to train PLMs, so that keeping the original name may be a simple but effective way to utilize the knowledge the model has learned. 
For single-relation subgraph, the serialization pattern is ``\textsf{<H> [head entity name] <D> [head entity description] <P> [predicate] <T> [tail entity name] <D> [tail entity description]}'', the special tokens <H>, <D>, <P>, and <T> mean the head entity, description, predicate and tail entity, respectively. 
For CVT subgraph, we add a special token ``\textsf{R@}'' in front of inside predicates (point to a CVT node) in order to traverse the subgraph in a uniform format. Therefore, the serialization pattern of one edge in the subgraph is ``\textsf{<P> [R@][predicate] <T> [entity name] <D> [entity description]}'', and then we just concatenate all edges together. Figure \ref{fig:ap_overview} gives an example.

\paragraph{Entity type placeholder} In this setup, we replace the entity name with its entity type in both input and target text. Intuitively, the delexicalization will make model focus more on predicate rephrasing.
Previous work \cite{elsahar_zero-shot_2018} picked the entity type that is mentioned most in the first sentence of the entity's Wikipedia article. However, it does not make sense to fix the entity type in all contexts.
Instead, we directly utilize the head and tail entity types contained in a Freebase predicate. The predicate in Freebase is organized as ``\textsf{domain.subject\_type.property}'', where we treat the ``\textsf{property}'' part as the type of object entity.
For example, the part of serialization shown in the Figure \ref{fig:ap_overview} a will be: ``\textsf{<P> R@film.film\_character. portrayed\_in\_films <T> [film\_character] <D> [film\_character] is the fictional character of 1985 film Return to Oz .}''.
In general, the serialization patterns used here are similar to those in the entity name setup.

\subsubsection{Collection of Subgraph Description}
Given a atomic subgraph $g$, the auto-prompter is trained to generate the corresponding subgraph description, denote as $\bar{d}$. 
Training the aforementioned auto-prompter needs large-scale labeled SPARQL and NL description pairs, since the SPARQL involves a lot of KB predicates, e.g.,  ``\textsf{film.film.actor}'', and SPARQL operators, e.g., ``\textsf{ORDER BY ?num DESC LIMIT 1}'' (means ``\textsf{Argmax}'').
For example, given a non-NL SPARQL (a simplified version of the Figure \ref{fig:ap_overview} b) ``\textsf{SELECT DISTINCT ?x WHERE \{m.01d1st film.actor.film ?y. ?y film.performance.film ?x .\}}'' and the name of topic entity \textsf{m.01d1st} ``Nick Cannon'', an annotator needs to understand the SPARQL and write the corresponding NL text ``\textsf{Nick Cannon star in film [?x]}''. 
Labelling such large-scale data is impracticable.
Therefore, the motivation here is to make use of large-scale unstructured corpus to fill the gap between predicates and natural language expression.
Different from \cite{elsahar_zero-shot_2018} who use a heuristic string matching rule to find the phrases in corpus by co-occurrence of entity names, we propose a novel soft-generation approach by combining the distant supervision and the ability of modern generative PLMs together. 

Specifically, to find the NL descriptions for atomic subgraphs, we rely on the Wikipedia 2018-12-20 dump as the source of text documents. 
Each page in Wikipedia has a title and a content that consists of a list of paragraphs. All these fields are re-tokenized by Spacy\footnote{\url{https://spacy.io}} and indexed by Elasticsearch\footnote{\url{https://www.elastic.co/elasticsearch}}.
As mentioned in \cite{riedel_modeling_2010, elsahar_zero-shot_2018}, we believe that the distant supervision assumption has been effective on Wikipedia. For single-relation subgraph $g$, we match sentence $\bar{d}$ in Wikipedia if the subject name and the object name of this triple co-occur in the same sentence.
For CVT subgraph $g$, we find paragraphs $\bar{d}$ in Wikipedia if all entities of this subgraph co-occur in the same paragraph. 
In addition, we remove the sentences that do not have any entity in matched paragraphs and drop subgraphs that match nothing. 
Furthermore, during the training phase, the entity names in the descriptions are replaced by their types in the entity type placeholder setup, while in another setup, the descriptions remain untouched. 
Accordingly, we need to replace the entity types with the corresponding names in the inference phase to get the prompt text.

\subsubsection{Training}

The auto-prompter is based on an advanced generative PLM, BART\cite{lewis_bart_2020}.
Specifically, in order to make the optimization process more stable, we merge and shuffle the two kinds of subgraph as training data, making them evenly distributed in each mini-batch. Then we fine-tune two auto-prompters based on the two serialization strategies mentioned above. The details of the hyper-parameters setting are recorded in Appendix \ref{hyper_setting}.
The auto-prompter is trained with a maximum likelihood objective. Given the training samples $(\bar{d}, g)$, the objective $\mathcal{L}$ is defined as:
\begin{equation}
	\mathcal{L} = - \sum_{M} \sum_{i=1}^{|\bar{d}|} \log p_{\Theta} (\bar{d}_i|\bar{d}_{<i}, g)
\end{equation}
where $\bar{d}_i$ is the $i$-th token in $\bar{d}$ and the $M$ is the length of training instances.

\subsubsection{Prompt Text}

The motivation for developing the prompt text is to smooth the transformation from non-NL SPARQL to NL question. 
Therefore we rephrase the formal expression in KB to a NL form with the help of PLMs. 
As shown in Figure \ref{fig:ap_overview} b, the auto-prompter successfully and correctly generates the relation, ``\textsf{A Very Scholl Gyrls Holla-Day starring Nick Cannon}'', from a relative complex CVT subgraph. 
In a word, given a SPARQL expression $S$, Section \ref{subgraph_sampling} instantiates and samples one RDF subgraph $G$.
Section \ref{subgraph_serialization} adds entity descriptions and serializes each $g \in G$.
Lastly, the well-trained auto-prompter is used to generate prompt text $T$ given $G$. 
It is hard to evaluate the quality of prompt text by automatic metrics because there is no target text. 
Therefore we conduct human evaluations and report the scores in Section \ref{human_eval}.

\subsection{Question Generation}\label{sec:qg}
\begin{figure}[t]
	\centering
	\includegraphics[width=0.48\textwidth,page=3]{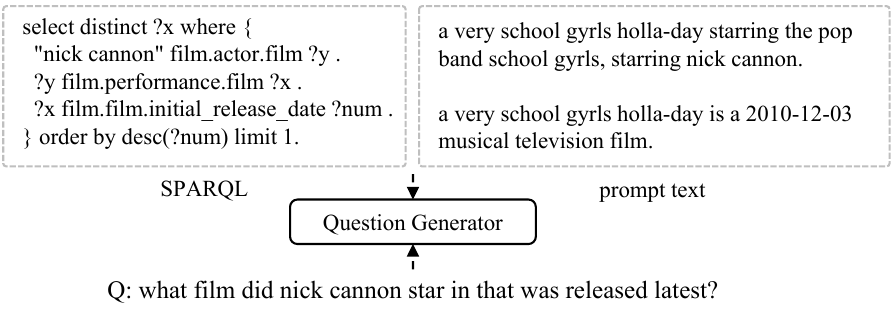}
	\caption{Overview of question generation.}
	\label{fig:qg}
\end{figure}
Question generation $\mathtt{QG}(\cdot)$ takes both the SPARQL $S$ and prompt text $T$ as input, and generates the question $\hat{Q}$ with max likelihood logit in the beam search space.
The question generator is implemented based on another BART. 
The training and inference processes here are similar to the classic fine-tuning methods of BART. 
The overall process is shown in Figure \ref{fig:qg}.

\subsubsection{Data Construction}

We construct a training sample by concatenating $S$ and $T$ together as input and the annotated question as target. 
In order to better prompt the question generator, we add ``\textsf{(the [var])}'' behind the corresponding instantiated entity in prompt text for each variable in SPARQL \footnote{For example, add ``\textsf{(the ?x)}''  behind ``\textsf{A Very School Gyrls Holla-Day}''}.

It is worth noting that, following previous works, we convert both input and output to lowercase.

\subsubsection{Training and Inference}

We fine-tune the question generator as a standard sequence-to-sequence model from the input to the output text, as presented in BART \cite{lewis_bart_2020}. Formally, given the training sample $(S, T, \bar{Q})$, the objective $\mathcal{L'}$ is defined as
\begin{equation}
	\mathcal{L'} = \sum_{N} p_{\Phi}(\bar{Q}|S, T)= - \sum_{N} \sum_{i=1}^{|\bar{Q}|} \log p_{\Phi} (\bar{Q}_i|\bar{Q}_{<i}, S, T)
\end{equation}
where $\bar{Q}_i$ is the $i$-th token in the target question $\bar{Q}$, $N$ is the length of training samples and $\Phi$ denotes the parameters of the question generator.

We use the standard beam search method for decoding. The specific parameter settings are reported in Appendix \ref{hyper_setting}.

%% file: 04-experiments.tex
\section{Experiments}

In this section, we conduct extensive experiments to evaluate the effectiveness of our proposed approach.

\subsection{Dataset \& Preprocessing}\label{dataset_preprocess}

\begin{table}
\caption{Statistics of the datasets.}
\label{tab:data_statistics}
\scalebox{0.85}{
  \begin{tabular}{lcccc}
  \toprule
  Dataset & \#Rel & \begin{tabular}[c]{@{}c@{}}\#Instances\\      (Train/Valid/Test)\end{tabular} & Total & Length \\ \hline
  WQCWQ1.1       & 931 & 29,909 / 2,529 / 2,529 & 34,967           & 14.12 \\
  \quad w/ OPs & 494 & 4,097 / 300 / 343      & 4,740 (13.56\%)  & 14.69 \\
  \quad w/o OPs   & 870 & 25,812 / 2,229 / 2,186 & 30,277 (86.44\%) & 14.03 \\
  PathQuestions   & 378 & 9,793 / 1,000 / 1,000  & 11,793           & 12.4  \\ \bottomrule
  \end{tabular}
}
\end{table}

\paragraph{WQCWQ1.1} WebQuestionsSP \cite{yih_value_2016} and ComplexWebQuestions 1.1 \cite{talmor_web_2018} are widely used question answering datasets that contain natural language questions and corresponding SPARQL queries. Previous works \cite{kumar_difficulty-controllable_2019, ke_jointgt_2021, chen_toward_2022} combined WebQuestionsSP and ComplexWebQuestions (the older version) and convert the SPARQL from \textsf{SELECT} query to \textsf{CONSTRUCT} query to get RDF graphs. However, ComplexWebQuestions is unavailable now, and previous works only released the processed results without original SPARQL expressions, wherein the exact semantics are lost. Therefore, we merge and reprocess these two datasets by ourselves and release a new dataset, WQCWQ1.1. We leave the original training set untouched and randomly divide the validation/test set equally.

\paragraph{PQ} PathQuestions (PQ) is constructed from a question answering dataset \cite{zhou_interpretable_2018}. The main characteristic is that PQ only consists of chain questions, wherein the path between the topic entity and the answer is 2-hop or 3-hop. Compared with WQCWQ1.1, PQ does not contain any complex operation in questions. We used the same data as the existing works \cite{kumar_difficulty-controllable_2019, chen_toward_2022}. 

Brief statistics of the datasets are listed in Table \ref{tab:data_statistics}, including the total number of relations, the data split, the subset divided by whether containing complex operations (OPs), and the average length of questions.

\subsection{Implementation}

Our auto-prompter and question generator are both based on pre-trained models BART \cite{lewis_bart_2020}. 
We initialize our model weights with the BART-base checkpoint released by HuggingFace's Transformers \cite{wolf_transformers_2020}.
We follow BART to use BytePair Encoding (BPE) vocabulary \cite{radford2019language} with the size of 50,265 and adopt Adam \cite{kingma_adam_2015} as the optimizer.
Since the computational resources are limited, we pick a set of hyper-parameters and train the auto-prompter on unsupervised data for 10 epochs. It took 60 hours on 4 NVIDIA A100 (40GB) GPUs. 
For training question generator, we apply different hyper-parameters on each data proportion setting. 
More details, including the hyper-parameters and search space settings, are reported in Appendix \ref{hyper_setting}.

\subsection{Baseline Methods}

We choose the following two categories of models as our baselines:

\paragraph{Pre-trained Models} We chose JointGT as the pre-trained baseline. JointGT \cite{ke_jointgt_2021} is a BART-based model for KG-to-text generation. It adopts a structure-aware semantic aggregation module to model the structure of an input graph at each Transformer layer. Afterward, it is pre-trained in large-scale KG-to-text corpora, and then fine-tuned in the downstream tasks, including question generation.

\paragraph{Task-Specific Models without Pre-training} We also adopted the recent task-specific models without pre-training as baselines, including Graph2Seq \cite{chen_toward_2022} and KTG \cite{bi_knowledge-enriched_2020}. Graph2Seq introduces a bidirectional graph encoder to model subgraphs in KG and generate questions by a graph-to-sequence generator with a coping mechanism. KTG proposes a knowledge-enriched, type-constrained, and grammar-guided model with auxiliary information to enrich input.

We report the baseline results directly if they use the same dataset as ours. Otherwise, we implement these baselines using the codes and parameters released by the original papers.

\subsection{Automatic Evaluation Metrics}

Following previous QG works \cite{kumar_difficulty-controllable_2019, bi_knowledge-enriched_2020, ke_jointgt_2021, chen_toward_2022} , we use BLEU-4 (B-4) \cite{papineni_bleu_2001}, METEOR (ME) \cite{denkowski_meteor_2014} and ROUGE-L (R-L) \cite{lin_rouge_2004} as our evaluation metrics. Initially, BLEU-4 and METEOR were designed to evaluate machine translation systems, and ROUGE-L was designed to assess text summarization systems.

\subsection{Experimental Results}

\begin{table*}
\centering
\caption{Results on WQCWQ1.1 and PathQuestions in six data proportion settings. The results marked with $\dagger$, $\ddagger$ and $\sharp$ are re-printed from the references \cite{bi_knowledge-enriched_2020}, \cite{chen_toward_2022} and \cite{ke_jointgt_2021}, respectively.}
\label{tab:main_res}
\scalebox{0.80}{
  \begin{tabular}{c | ccc | ccc | ccc | ccc | ccc | ccc}
  \toprule
  \multicolumn{19}{c}{WQCWQ1.1} \\
  \hline
  Data Proportion & \multicolumn{3}{c|}{0.1\%} & \multicolumn{3}{c|}{0.5\%} & \multicolumn{3}{c|}{1\%} & \multicolumn{3}{c|}{5\%} & \multicolumn{3}{c|}{10\%} & \multicolumn{3}{c}{100\%} \\ \hline
  Model & B-4 & ME & R-L & B-4 & ME & R-L & B-4 & ME & R-L & B-4 & ME & R-L & B-4 & ME & R-L & B-4 & ME & R-L \\ \hline
  Graph2Seq & 0.00 & 0.87 & 5.98 & 7.01 & 12.65 & 26.86 & 8.53 & 13.51 & 31.70 & 17.17 & 21.56 & 45.11 & 19.69 & 23.25 & 46.64 & 29.56 & 31.14 & 58.34 \\
  JointGT & 6.92 & 18.77 & 32.14 & 11.31 & 23.10 & 38.62 & 13.83 & 24.90 & 42.52 & 20.79 & 28.41 & 49.67 & 25.25 & 30.50 & 54.05 & 32.81 & 34.48 & 60.92 \\ \hline
  AutoQGS & \textbf{15.26} & \textbf{22.07} & \textbf{42.97} & \textbf{21.56} & \textbf{27.06} & \textbf{48.70} & \textbf{24.09} & \textbf{28.77} & \textbf{51.04} & \textbf{29.58} & \textbf{32.43} & \textbf{56.90} & 31.81 & \textbf{33.82} & 59.07 & \textbf{36.93} & \textbf{36.63} & \textbf{63.82} \\
  AutoQGS-T & 14.43 & 21.45 & 42.59 & 20.40 & 26.09 & 47.53 & 23.18 & 28.11 & 50.40 & 28.86 & 31.83 & 56.13 & \textbf{32.00} & 33.79 & \textbf{59.52} & 36.49 & 36.38 & 63.53 \\
  \hline \hline
  \multicolumn{19}{c}{PathQuestions} \\
  \hline
  KTG & \multicolumn{3}{c|}{-} & \multicolumn{3}{c|}{-} & \multicolumn{3}{c|}{-} & \multicolumn{3}{c|}{-} & \multicolumn{3}{c|}{-} & 45.58$\dagger$ & 52.31$\dagger$ & 73.21$\dagger$ \\
  Graph2Seq & 1.01 & 4.99 & 12.07 & 2.63 & 10.64 & 41.45 & 17.59 & 18.35 & 51.44 & 43.43 & 31.34 & 67.51 & 42.72 & 32.20 & 67.62 & 61.48$\ddagger$ & 44.57$\ddagger$ & 77.72$\ddagger$ \\
  JointGT & 43.15 & \textbf{35.91} & \textbf{69.57} & 51.05 & 41.23 & 73.23 & 51.89 & 42.19 & 73.62 & 55.90 & 43.25 & 74.49 & 57.39 & 43.51 & 75.26 & \textbf{65.89}$\sharp$ & \textbf{48.25}$\sharp$ & \textbf{78.87}$\sharp$ \\ \hline
  AutoQGS & \textbf{43.46} & 33.55 & 68.23 & \textbf{56.30} & \textbf{41.95} & \textbf{74.68} & \textbf{58.69} & \textbf{42.48} & \textbf{75.50} & \textbf{61.55} & \textbf{44.81} & \textbf{76.68} & \textbf{60.73} & \textbf{44.92} & \textbf{76.95} & 65.13 & 47.50 & 76.80 \\
  \bottomrule
  \end{tabular}
  }
\end{table*}

\begin{table}
\caption{Average gains on WQCWQ1.1 and PathQuestions over six data proportion settings.}
\label{tab:avg_gain}
  \begin{tabular}{l|ccc|ccc}
  \toprule
  Dataset   & \multicolumn{3}{c|}{WQCWQ1.1} & \multicolumn{3}{c}{PathQuestions} \\ \hline
  Metrics   & B-4      & ME       & R-L     & B-4       & ME        & R-L       \\ \hline
  Graph2Seq & 12.88    & 12.97    & 17.98   & 29.82     & 18.83     & 22.26     \\
  JointGT   & 8.06     & 3.44     & 7.43    & 3.43      & 0.15      & 0.63      \\
  \bottomrule
  \end{tabular}
\end{table}

\subsubsection{Automatic Evaluation}
Table \ref{tab:main_res} shows the detailed evaluation results comparing our proposed models against other state-of-the-art baselines in six data proportion settings, from 0.1\% to 100\%, respectively. 
AutoQGS is implemented on the entity name serialization setup by default, and we also report the experimental results on entity type placeholder serialization setup on WQCWQ1.1, denoted as AutoQGS-T.
Since AutoQGS performs better than AutoQGS-T under most settings on WQCWQ1.1, we only report AutoQGS performance hereafter.
As we can see, both AutoQGS and AutoQGS-T outperform every baseline by a large margin on WQCWQ1.1, particularly in few-shot settings.
Table \ref{tab:avg_gain} shows the mean average gains across six settings. AutoQGS generally exceeds Graph2Seq/JointGT by 12.88/8.06 BLEU-4 points in WQCWQ1.1, and 29.82/3.43 in PQ, respectively. 
Specifically, for Graph2Seq, the vocab depends on the training data heavily. 
As the training instances decrease, the vocabulary becomes smaller and more words become OOV (out of vocabulary), which severely degrades the performance of this model.
Furthermore, our model outperforms JointGT in all six settings except in 0.1\% of PQ. 
We speculate that, in the 0.1\% setting, there are only ten training instances (9793*0.001, we take the upper bound), which is too difficult for both JointGT and AutoQGS. 
These results verify that AutoQGS can effectively and accurately generate questions based on SPARQL in low-resource scenarios.

\subsubsection{Impact of Complex Operations}

\begin{table}
\centering
\caption{Results on WQCWQ1.1 subsets dividing by whether containing complex operations.}
\label{tab:impact_of_op}
  \begin{tabular}{c | ccc | ccc}
  \toprule
  WQCWQ1.1 & \multicolumn{3}{c|}{w/ OPs}  & \multicolumn{3}{c}{w/o OPs} \\ \hline
  Model & B-4 & ME & R-L & B-4 & ME & R-L \\ \hline
  Graph2Seq & 22.17 & 26.63 & 51.72 & 30.11 & 31.32 & 58.79 \\
  JointGT & 24.33 & 30.83 & 54.35 & 35.08 & 35.35 & 62.48 \\
  AutoQGS & \textbf{35.75} & \textbf{36.73} & \textbf{62.11} & \textbf{36.03} & \textbf{36.09} & \textbf{63.37} \\
  \bottomrule
  \end{tabular}
\end{table}

Next, we evaluate how the complex operations affect the performance of our AutoQGS.
We keep the same divisions in WQCWQ1.1 and split data into two subsets by whether containing complex operations.
The statistics are reported in Table \ref{tab:data_statistics}. About 86.59\% of the data do not contain any complex operation, which means that this part is similar to the data used by previous works. 
We report the comparison results on both subsets in Table \ref{tab:impact_of_op}.
Results show that AutoQGS performs better in both settings. 
The results confirm the point that our model can generate questions from SPARQL better than others.

\subsection{Ablation Test}

\begin{table}
\caption{Ablation tests on WQCWQ1.1}
\label{tab:ablation_test}
\centering
\scalebox{0.95}{
  \begin{tabular}{l|lll|lll}
  \toprule
  Data Proportion                & \multicolumn{3}{c|}{0.1\%} & \multicolumn{3}{c}{1\%}  \\ \hline
  Model & B-4 & \multicolumn{1}{c}{ME} & \multicolumn{1}{c|}{R-L} & \multicolumn{1}{c}{B-4} & \multicolumn{1}{c}{ME} & \multicolumn{1}{c}{R-L} \\ \hline
  AutoQGS                      & 15.26   & 22.07   & 42.97   & 24.09   & 28.77  & 51.04 \\
  \quad w/o prompt text          & 13.72   & 20.87   & 42.46   & 18.69   & 24.39  & 46.75 \\
  \quad re/desc                  & 11.65   & 19.16   & 40.39   & 21.92   & 26.66  & 48.09 \\ \hline
                                 & \multicolumn{3}{c|}{10\%}   & \multicolumn{3}{c}{100\%} \\ \hline
  AutoQGS                      & 31.81   & 33.82   & 59.07   & 36.93   & 36.63  & 63.82 \\
  \quad w/o prompt text          & 31.07   & 33.08   & 58.79   & 36.18   & 36.19  & 63.57 \\
  \quad re/desc                  & 31.49   & 33.31   & 58.82   & 36.00   & 36.17  & 63.19 \\
  \bottomrule
  \end{tabular}
}
\end{table}

We conduct ablation tests to investigate the effectiveness of AutoQGS in four representative data proportion settings (0.1\%, 1\%, 10\%, 100\%) by removing the prompt text and replacing the prompt text with topic entity descriptions (\textsf{re/desc}, for short) one at a time. 
As shown in Table \ref{tab:ablation_test}, removing the prompt text leads to significant performance reduction in all settings, particularly in 0.1\% and 1\%. The decline is consistent with our purpose as the prompt text is designed to enable the model to perform better in low-resource scenarios.
Similarly, we also observe performance decline by replacing prompt text with topic entity descriptions. This result again validates the effectiveness of prompt text, as it is also developed to mine relations between entities rather than only describe them.

\subsection{Human Evaluation}\label{human_eval}

\begin{table}
\caption{Human evaluations results ($\pm$ standard deviation). Pred and Natural mean the percent of predicates identification and naturalness score (0-5), respectively.}
\label{tab:human_evaluation_both}
\scalebox{0.96}{
  \begin{tabular}{lcccc}
  \toprule
  \multicolumn{5}{c}{auto-prompter}                                                                                    \\ \hline
  \multicolumn{1}{l|}{Graph Type}      & \multicolumn{2}{c|}{Pred}                     & \multicolumn{2}{c}{Natural} \\ \hline
  \multicolumn{1}{l|}{single-relation}   & \multicolumn{2}{c|}{81.85\% (5.4)}                  & \multicolumn{2}{c}{4.25 (0.18)}   \\
  \multicolumn{1}{c|}{CVT}               & \multicolumn{2}{c|}{71.11\% (7.2)}                  & \multicolumn{2}{c}{3.66 (0.25)}   \\ \hline \hline
  \multicolumn{5}{c}{question generator}                                                                               \\ \hline
  \multicolumn{1}{l|}{Data proportion} & \multicolumn{2}{c|}{0.1\%}                    & \multicolumn{2}{c}{1\%}     \\ \hline
  \multicolumn{1}{l|}{Model}             & Pred       & \multicolumn{1}{c|}{Natural}     & Pred         & Natural      \\ \hline
  \multicolumn{1}{l|}{JointGT}           & 44\% (4.2) & \multicolumn{1}{c|}{2.76 (0.16)} & 72\% (4.9)   & 3.72 (0.22)  \\
  \multicolumn{1}{l|}{AutoQGS}           & 56\% (6.5) & \multicolumn{1}{c|}{4.36 (0.19)} & 80\% (5.0)   & 4.60 (0.24)   \\ \hline
  \multicolumn{1}{l|}{}                  & \multicolumn{2}{c|}{10\%}                     & \multicolumn{2}{c}{100\%}   \\ \hline
  \multicolumn{1}{l|}{JointGT}           & 80\% (3.5) & \multicolumn{1}{c|}{4.20 (0.30)}  & 84\% (3.3)   & 4.28 (0.19)  \\
  \multicolumn{1}{l|}{AutoQGS}           & 84\% (4.1) & \multicolumn{1}{c|}{4.68 (0.32)} & 88\% (3.9)   & 4.84 (0.16)  \\ \hline \hline
  \multicolumn{1}{c|}{}                  & \multicolumn{2}{c|}{Pred}                     & \multicolumn{2}{c}{Natural} \\ \hline
  \multicolumn{1}{l|}{Golden}            & \multicolumn{2}{c|}{95\% (2.1)}                  & \multicolumn{2}{c}{4.80 (0.18)}   \\ \bottomrule
  \end{tabular}
}
\end{table}

To further evaluate AutoQGS, we conduct two human evaluations on the auto-prompter and question generator results, respectively. 
Considering that the goals of the two components are essentially similar, we decide to run the same two criteria following  \cite{elsahar_zero-shot_2018}.

\paragraph{Predicates identification} Annotators were asked to estimate whether the generated text expresses all predicates in the given SPARQL (or subgraph) or not. 
\paragraph{Naturalness} Annotators were requested to assign each generated text based on the fluency and readability by a score from 1 to 5, where (5) perfectly clear and natural, (3) grammatically correct but seems artificial, and (1) entirely not understandable. 

For the evaluation of auto-prompter, we randomly sample 100 atomic subgraphs from the dataset. 
For the evaluation of question generator, we randomly sample 100 instances from the test set.
We also collect the outputs from the most competitive baseline, JointGT, for comparison. 
All evaluations are done with the help of 3 annotators. 
Results in Table \ref{tab:human_evaluation_both} show some critical observations. 
First of all, the auto-prompter has the ability to paraphrase predicates consistently and fluently. On the other hand, the question generator achieves remarkable results in all four settings, whereas the naturalness score in the 100\% setting is even higher than that of annotated questions (denoted as Golden).
Generally speaking, our AutoQGS can beat the corresponding baselines in both predicates identification and naturalness.

\subsection{Case Study}

\begin{figure*}
  \centering
  \includegraphics[width=1\textwidth,page=4]{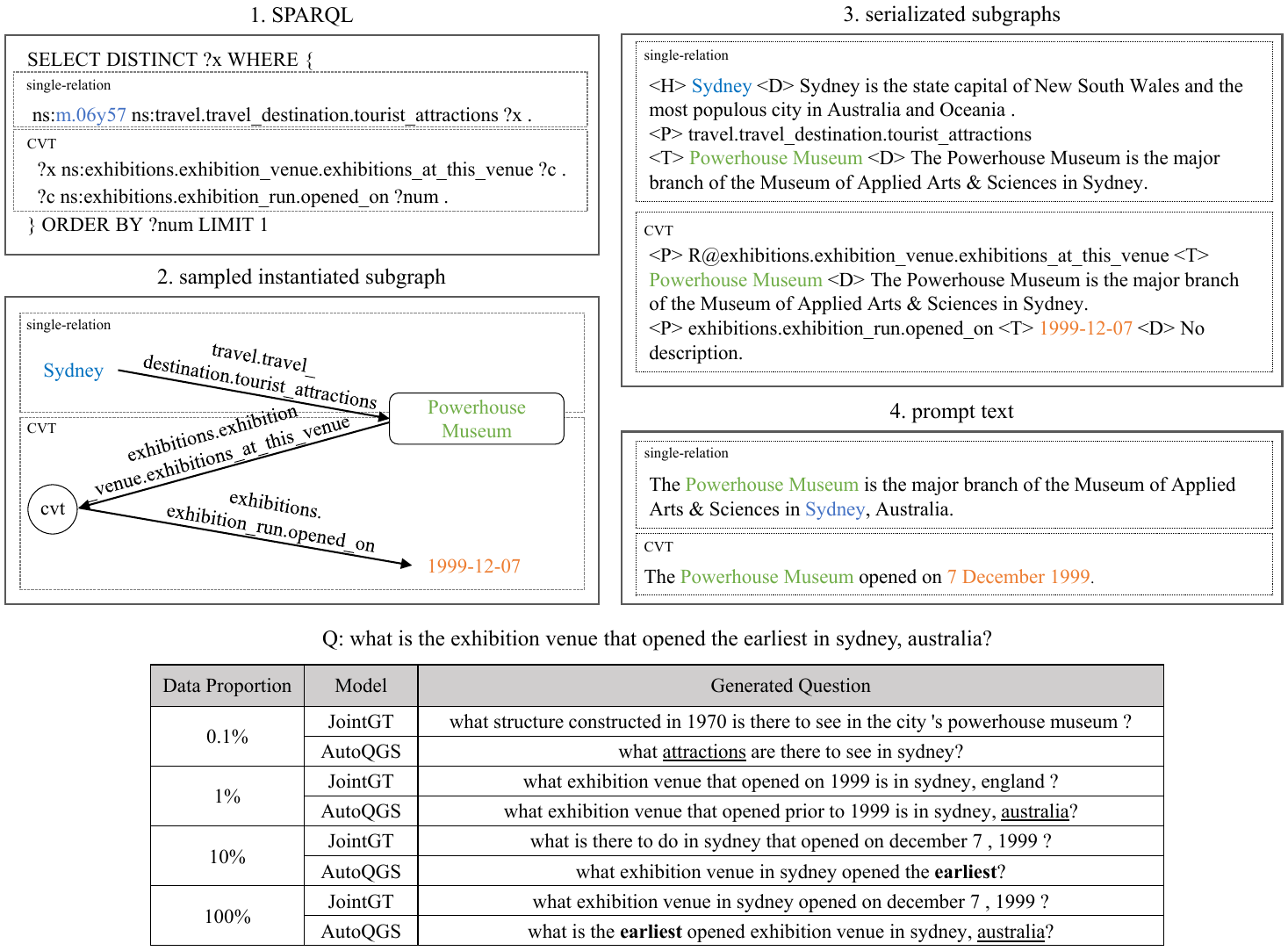}
  \caption{Case study}
  \label{fig:case}
\end{figure*}

To provide a complete and visual presentation of AutoQGS, we provide a case in Figure \ref{fig:case}.
Steps 1 to 4 display the processes of how to get the prompt text by the auto-prompter. 
Firstly, we re-write the SPARQL query (adding ``\textsf{?c ?num}'' in this case) to search all variables in KG and sample one from the results to instantiate a complete subgraph. 
Secondly, for each atomic subgraph in the complete subgraph, we serialize it in corresponding pattern.
Afterward, we use the well-trained auto-prompter to get the prompt text of every atomic subgraph. 
Finally, in order to explicitly describe the relationship between entities and variables, we add ``\textsf{(the [var])}'' behind the corresponding entities in the prompt text, as shown in Step 4.
The table in Figure \ref{fig:case} demonstrates the quality of generated questions (converted to lowercase). 
Compared to JointGT, AutoQGS can generate questions more faithfully and completely. 
More importantly, AutoQGS is able to generate questions with complex operations according to the SPARQL accurately.
For example, JointGT fails to express ``\textsf{earliest}'' for ``\textsf{ORDER By ?num LIMIT 1}'' expressed in SPARQL in all settings, whereas ours successfully capture it with only 10\% training data (marked in bold). 
On the other hand, our model successfully generate ``\textsf{Australia}'' with the help of prompt text (marked in underscore). It proves that the prompt text successfully supplements information for question generation. 
Moreover, in 0.1\% setting, the question generated by ours is much more natural and fluent. 

\subsection{Error Analysis}

\begin{table}
\caption{Error analysis}
\label{tab:error_analysis}
  \begin{tabular}{l|p{6cm}}
  \toprule
  SPARQL &
    \begin{tabular}[p{6cm}]{@{}l@{}}SELECT DISTINCT ?x WHERE \{\\ \quad "Julius Caesar" people.deceased\_person.place\\  \_of\_death "The Theatre of Pompey"(the ?x) .\\ \}\end{tabular} \\ \hline
  Prompt text &
    The Theatre of Pompey (the ?x) was built during the reign of Julius Caesar. \\ \hline
  Golden &
    where was caesar when he \textbf{was stabbed}? \\ \hline
  AutoQGS &
    where did julius caesar die? \\ \hline \hline
  SPARQL &
    \begin{tabular}[p{6cm}]{@{}l@{}}SELECT   DISTINCT ?x WHERE \{\\ \quad "Chile" location.country.form\_of\_government \\ "Presidential system"(the ?x) .\\  \quad "Presidential system"(the ?x) government.form\\  \_of\_government.countries "Brazil" .\\ \}\end{tabular} \\ \hline
  Prompt text &
    Chile has a Presidential system (the ?x) with a bicameral legislature. Brazil is a country with a Presidential system (the ?x) \\ \hline
  Golden &
    what are the government types of chile and brazil? \\ \hline
  AutoQGS &
    what type of government \textbf{is used} in both brazil and chile? \\
  \bottomrule
  \end{tabular}
\end{table}

Table \ref{tab:error_analysis} shows some failure cases on the WQCWQ1.1 test set in 100\% data proportion setting. 
For convenience, variables in SPARQL are instantiated and put together with their variable names.
The most common mistake for the auto-prompter is that the distant supervision approach sometimes produces descriptions that are not consistent with the facts, as shown in the first case\footnote{In this example, the Theatre of Pompey was built during the latter part of the Roman Republican era by Pompey the Great, not Julius Caesar.}. 
On the other hand, one of the frequent error patterns we find out for the question generator is incorrect statement. 
For instance, in the second example, the statement ``\textsf{what type of government is used}'' is grammatically correct, but the usage is inappropriate. 
Another error pattern is lack of knowledge. 
In the first example, given the SPARQL and the prompt text, we still don't know that Caesar was stabbed. 
But it is usual to include common sense in a question.

\subsection{Data Augmentation}
Automatically constructing question-answer pairs from knowledge bases is one of the objectives of question generation. 
Based on the WQCWQ1.1 dataset, we augment data by replacing topic entities in SPARQL. 
Specifically, given a SPARQL, we construct a new one to get a list of different topic entities by replacing the topic entities with variables.
For each set of entities in the result, we construct another SPARQL, in which only the named entity is different from the original one, to form a new sample.
Afterward, we use the best AutoQGS trained in 100\% data to generate the corresponding questions. 
In order to multiply the data tenfold, we randomly choose ten instances from querying results for each item in WQCWQ1.1.
Finally, we generate a corpus of 340K question-SPARQL pairs for future research on question generation and question answering.

%% file: 05-conclusion.tex
\section{Conclusion}
In this paper, we propose AutoQGS, an auto-prompt approach for low-resource KBQG from SPARQL.
Firstly, we generate questions directly from SPARQL to handle complex operations. 
Secondly, we propose an auto-prompter to rephrase SPARQL into NL prompt text, smoothing the transformation from non-NL SPARQL to NL question with PLMs.
Lastly, we devise a question generator to generate questions given SPARQL and corresponding prompt text. 
Experimental results show that our approach achieves state-of-the-art performance especially in low-resource scenarios.
Furthermore, we generate a dataset of 330k factoid complex question-SPARQL pairs for further KBQG research.

%% file: 06-acks.tex
\begin{acks}
This work was supported by the National Key Research and Development Program of China under Grant No. 2020AAA0108600 and No. 2020YFC0833301.
\end{acks}

%% file: 07-appendix.tex

\appendix

\section{Appendix}

\subsection{Hyper-Parameter Setting}\label{hyper_setting}

\begin{table}[htp]
\caption{Hyper-parameter search space. \textsf{[]} indicates that the listed numbers will be traverse one by one.  }
\label{tab:hyper_search}
    \begin{tabular}{l|c|c}
    \toprule
    Model                 & auto-prompter & question generator     \\ \hline
    Hyper-parameter       & Setting       & Search Space           \\ \hline
    Learning Rate         & 3e-5          & {[}3e-5, 5e-5, 1e-4{]} \\
    Warmup Proportion     & 0.1           & {[}0.1, 0.2, 0.3{]}    \\
    Batch Size            & 64            & {[}16, 24, 32, 64{]}   \\
    Beam Size             & 10            & {[}5, 10{]}            \\
    Length Penalty        & -             & {[}1.0, 1.2{]}         \\
    Input Length          & 512           & 512                    \\
    Output Length         & 512           & 128                    \\
    Warmup Epoch          & 10            & 50                     \\
    Early Stop Patience   & -             & 10                     \\
    Maximum Gradient Norm & 1.0           & 1.0                    \\
    Optimizer             & Adam          & Adam                   \\
    Epsilon (for Adam)    & 1e-8          & 1e-8                   \\ 
    \bottomrule
    \end{tabular}
\end{table}

\begin{table}[htp]
\caption{Best assignments of hyper-parameters for question generator.}
\label{tab:hyper_assignments}
    \begin{tabular}{c|cccccc}
    \toprule
    Dataset           & \multicolumn{6}{c}{WQCWQ 1.1}            \\ \hline
    Data Proportion   & 0.1\% & 0.5\% & 1\% & 5\% & 10\% & 100\% \\ \hline
    Warmup Proportion & 0.2   & 0.1   & 0.1 & 0.2 & 0.2  & 0.2   \\
    Batch Size        & 16    & 16    & 32  & 32  & 32   & 64    \\
    Beam Size         & 10    & 10    & 10  & 10  & 10   & 10    \\
    Length Penalty    & 1.2   & 1.0   & 1.2 & 1.0 & 1.0  & 1.0   \\ \hline \hline
    Dataset           & \multicolumn{6}{c}{Pathquestions}        \\ \hline
    Data Proportion   & 0.1\% & 0.5\% & 1\% & 5\% & 10\% & 100\% \\ \hline
    Warmup Proportion & 0.1   & 0.1   & 0.1 & 0.1 & 0.1  & 0.1   \\
    Batch Size        & 16    & 16    & 16  & 24  & 24   & 24    \\
    Beam Size         & 10    & 10    & 10  & 5   & 5    & 5     \\
    Length Penalty    & 1.2   & 1.2   & 1.2 & 1.0 & 1.0  & 1.0   \\ 
    \bottomrule
    \end{tabular}
\end{table}

We provided the detailed settings of hyper-parameters for training the auto-prompter and the question generator in Table \ref{tab:hyper_search}.
The table include the hyper-parameter settings for training auto-prompter and the hyper-parameter search space for training question generator.
We implement our models base on Huggingface’s Transformers \cite{wolf_transformers_2020}.
For auto-prompter, we train the model on unsupervised data for 10 epochs.
For question generator, We set \textsf{Warmup Epoch}, which means the expected training rounds, to control warmup steps.
The hyper-parameter search space and the best assignments are listed in Table \ref{tab:hyper_search}. We go through each combination of hyper-parameters in every few-shot settings to find the optimal one. The selection criterion is BLEU-4 on the validation set.